# Comparative Study of Out-of-the-Box Technology for Automatic Target Detection and Recognition

Alma M. Liezenga[a], Lotte Nijskens[a], Henrik R. Baumann[b], Stefan Becker[c], Simon Bensberg[d], Niccolò Camarlinghi[e], Håvard R. Eiring[b], Alexander W. Johnsgaard[b], Tanel Liiv[f], Giuseppe Martino[e], Matteo Marturini[g], Matthias Rapp[h], Jan Erik van Woerden[a], Alexander Wolpert[h] and Hugo J. Kuijf[a]

[a] *Intelligent Imaging, TNO Defence, Safety and Security,* The Hague, The Netherlands
{alma.liezenga, lotte.nijskens, jan_erik.vanwoerden, hugo.kuijf}@tno.nl
[b] *Division Defence Systems, Kongsberg Defence & Aerospace,* Kongsberg, Norway
{henrik.rambech.baumann, havard.eiring, alexander.waller.johnsgaard}@kongsberg.com
[c] *Object Recognition OBJ, Fraunhofer IOSB,* Karlsruhe, Germany
stefan.becker@iosb.fraunhofer.de
[d] *Artificial Intelligence & Computer Vision, Rheinmetall Electronics GmbH,* Bremen, Germany
simon.bensberg@rheinmetall.com
[e] *FlySight S.r.l.*, Livorno, Italy
{niccolo.camarlinghi, giuseppe.martino}@flysight.it
[f] *Marduk Technologies OÜ,* Tallinn, Estonia
tanel.liiv@marduk.ee
[g] *Digital Safety & Security, Data Science & Artificial Intelligence, Austrian Institute of Technology GmbH,* Vienna, Austria
matteo.marturini@ait.ac.at
[h] *Hensoldt Optronics GmbH,* Oberkochen, Germany
{matthias.rapp, alexander.wolpert}@hensoldt.net

*Abstract*— **Automatic Target Detection and Recognition (ATD/R) is critical for military decision support and (semi-) autonomous operations. Recent advances in object detection and artificial intelligence (AI) significantly boosted the potential performance of ATD/R. However, the scarcity of publicly available military datasets limits the application of these systems. As a solution, this paper explores the use of publicly available models and civilian datasets to achieve reasonable performance in military contexts. We benchmark several state-of-the-art models, including six iterations of the YOLO series and two variations on the DETR framework, on a newly acquired military relevant dataset. This dataset features military vehicles and challenging circumstances, including various degrees of occlusions and small targets. The out-of-the-box version of each model is validated alongside a version fine-tuned on the VisDrone dataset. This dataset features small objects, an Air-to-Ground (A2G) perspective and relevant classes, potentially generalizing to our military ATD/R task. We compare the performance of the models using mAP@0.5 and mAP@0.5:0.95, across A2G and Ground-to-Ground (G2G) perspective, target size and model size, giving insight into the real-time capabilities of models. Our main findings are: (1) bigger models outperform smaller models, (2) DETR-based models show promising results compared to the YOLO series, (3) fine-tuning models on an out-of-domain A2G dataset, improves their A2G performance and slightly improves their performance on small objects, but (4) all models still struggle with detecting small objects in an A2G scenario. We conclude that, despite recent advances in object detection, in-domain training is still crucial for creating capable ATD/R systems.**

This paper was originally presented at the International Conference on Military Communication and Information Systems (ICMCIS), organized by the Information Systems Technology (IST) Scientific and Technical Committee, IST-224-RSY – the ICMCIS, held in Bath, United Kingdom, 12-13 May 2026.

## I. Introduction

The use of artificial intelligence (AI) and deep learning techniques has enabled significant progress in object detection in images and videos [1]. Most current approaches rely on publicly available datasets and pretrained deep learning models [2]. Unfortunately, these models often struggle to generalize to out-of-domain data [3], meaning that models trained on civilian datasets may perform poorly in military use-cases. As a result, Automatic Target Detection and Recognition (ATD/R) of military vehicles remains a considerable challenge.

Potential solutions include acquiring more datasets from the military domain, applying zero-shot or few-shot machine learning strategies [4], and generating synthetic imagery [5, 6] to increase the size and diversity of existing training sets. Recent work has demonstrated the few- and zero-shot capabilities of publicly available techniques, including the DEtection TRansformer (DETR) framework [7]. In addition, newer versions of the YOLO series and DETR framework have shown strong performance in a wide range of applications [8, 9, 10]. These developments prompt us to revisit the task of military ATD/R using out-of-the-box techniques and investigate whether in-domain training and the acquisition of large-scale military datasets are still a necessity.

In this study, we leverage a newly acquired military dataset, collected as part of the EDF project Shared daTabase for Optronic Image Recognition and Evaluation (STORE), providing a unique opportunity for an operationally representative and objective quantitative evaluation. Since

European Defence Fund through the project STORE, grant agreement №101121405

this dataset is newly acquired and solely available to the consortium partners, it is guaranteed to not have been used in pretraining of any publicly available models. We evaluate the performance of six YOLO iterations and two DETR variants in both air-to-ground (A2G) and ground-to-ground (G2G) scenarios, across object sizes and model sizes for out-of-the-box performance on the military dataset. Additionally, all methods are fine-tuned on the publicly available VisDrone dataset [11] to measure performance gains in A2G scenarios.

## II. METHOD

This section describes the ATD/R task, datasets, models and experiments used to investigate to which degree publicly available datasets and models can be used for a military ATD/R task.

### *A. Automatic Target Detection and Recognition*

The key task in this work was ATD/R of both persons and vehicles in a military context. All models were evaluated in an A2G and G2G setting. Vehicles included both civilian vehicles (cars, buses) and military vehicles (tanks, trucks[1]). These were merged into one vehicle class, to facilitate evaluation of methods trained on a civilian dataset in the military domain. The evaluation was performed on a newly acquired, military relevant dataset that also included occlusion of and small objects. All models were prepared in two ways: (1) models were applied out-of-the-box (zero-shot) with a mapping of their original classes to two super-classes 'person' and 'vehicle', and (2) models were fine-tuned on the VisDrone dataset (a civilian A2G setting) [11].

### *B. Datasets*

Six datasets were used for training and evaluation in this research: (1) MS COCO [12], which almost all models were (pre)trained on, (2) Object365 [13], (3) GoldG [14] and (4) CC3M-Lite [15] , which some models were pretrained on, (5) VisDrone [11], which the models were fine-tuned on, and (6) the STORE dataset, which the models were evaluated on. An overview of which models used which dataset(s) for (pre)training can be found in TABLE I. though the exact split and processing of the dataset can vary depending on the exact implementation of the model developers.

**MS COCO** [12] (MicroSoft Common Object in COntext) is a dataset released to advance state-of-the-art object detection. The dataset contains 328,000 images with a total of 2.5 million annotations from 91 classes, such as car, person, apple, clock, and suitcase [12]. The dataset is routinely used for pretraining state-of-the-art models, including the models used in this study. The classes from MS COCO were mapped to classes relevant to our task. The following classes were mapped to 'vehicle': 'bicycle', 'car', 'motorcycle', 'bus', 'truck', 'train', 'boat', and 'airplane'. There was only one 'person' class in MS COCO.

**Object365** [13] is a large-scale, high-quality dataset for object detection. The dataset contains 600,000 training images with a total of 10 million annotations from 365 classes [13]. Though less popular than MS COCO, this dataset has also been used for pretraining of models included in this study. The following classes from Object365 were mapped to 'vehicle': 'Car', 'SUV', 'Van', 'Bus', 'Motorcycle', 'Bicycle', 'Truck', 'Pickup Truck', 'Tricycle', 'Fire truck', 'Heavy Truck', 'Ambulance', 'Machinery Vehicle', 'Train', 'Sports Car', 'Scooter', 'Rickshaw', 'Carriage', 'Crane', 'Helicopter', 'Airplane', 'Boat', 'Sailboat', 'Ship', 'Hot-air balloon', and 'Formula 1'. There was only one 'person' class in Object365.

**GoldG** [14] is a combination of Flickr30k [16], an image description dataset consisting of 31,000 images with 276,000 manually annotated bounding boxes with dense text descriptions, and GQA [17], a large-scale benchmark for real-world visual reasoning and compositional question answering. GQA contains around 113,000 images paired with millions of question–answer pairs, each supported by detailed scene graphs that list objects, their attributes, and relationships.

**CC3M-Lite** [15] is an image captioning dataset containing 3.3 million samples from a variety of settings, including natural images, product images, drawings etc. The dataset was created using a pipeline that processes billions of webpages in parallel, and produces a set of image-caption pairs. The captions are filtered and processed from the original Alt-text attributes of the HTML webpage.

**VisDrone** [11] is a large-scale visual object detection and tracking dataset captured using drones over various Chinese urban and suburban areas. The dataset contains 263 videos and 10,209 images with a total of 2.5 million annotations from ten classes [18, 11]. The dataset features a large degree of variety in terms of weather conditions, viewpoints, and altitude; and represents an A2G viewpoint. The dataset has been used in multiple challenges for evaluating object detector performance [18]. Examples of images from the VisDrone dataset are shown in Fig. 1.

The official training split of the VisDrone2019-DET dataset [19] was used for fine-tuning our models, and the official validation split was used to determine the best checkpoint during training. The dataset was preprocessed to take into account the official ignored regions: these areas were set to a fixed grey color (RGB values (144, 144, 144)). The remaining annotations that had significant overlap with the ignored regions (IoU>= 0.8) were removed. The following classes were mapped to 'vehicle': 'car', 'van', 'truck', and 'bus'. Labels for two- or three-wheeled vehicles were not considered in the final dataset ('bicycle', 'tricycle', 'awning-tricycle', and 'motor'). The 'pedestrian' class was mapped to 'person'.

**The STORE dataset** [20] is a non-public military dataset that was acquired on two occasions, (1) at the military site in Fontevraud-l'Abbaye, France and (2) at a testing facility near Kongsberg, Norway; as part of the EDF project STORE. The selected dataset contains 3,316 frames from 80 videos with a total 10,683 annotations from the 'person' (5,414) and 'vehicle' (5,269) classes. The vehicle class includes military[1] and civilian vehicles (cars and vans).

---

[1] T62 and T72 battle tanks, BMP1 tracked infantry fighting vehicle, MTLB and SISU XA (tracked/untracked) armored personnel carriers, 2S1 and 2S3 tracked armored self-propelled artillery howitzers and Mowag Eagle, wheeled armored vehicle were included in the recording campaigns.

The STORE dataset was recorded using a combination of drones, stationary, and moving ground vehicles, thus including videos from both an A2G and G2G perspective. The dataset was recorded to represent an operational, military context and thus includes varying conditions: one portion being recorded in Nordic, snowy conditions and the other portion in a milder Mid-European climate with some rainfall. Object are at times (partly) occluded by trees or rainfall.

Out of the 80 videos, 13 (654 frames) were recorded in Fontevraud and 67 (2,662 frames) in Kongsberg. The videos from Fontevraud were recorded with an A2G perspective and the videos from Kongsberg were recorded with a G2G perspective. The dataset was used solely for evaluation purposes in this study. A few examples of annotated images from the dataset are shown in Fig. 2 and Fig. 3.

### C. Models

A number of state-of-the-art object detection models were selected for this evaluation. Models were prepared in two ways: (1) the models were taken out-of-the-box without any fine-tuning and with mapping from their original classes to the relevant classes 'person' and 'vehicle' and (2) the models were fine-tuned on the VisDrone dataset using the default hyperparameters, unless stated otherwise. An overview of all included model versions, the datasets they were originally (pre)trained on, and the model sizes used in our experiments is provided in TABLE I. For all our experiments, the following standard hyperparameters were used: image size of 640x640, batch size of 16 and 8 (only for RT-DETR) and 100 training epochs with no early stopping.

Fig. 1. Two example images from the VisDrone dataset [19]

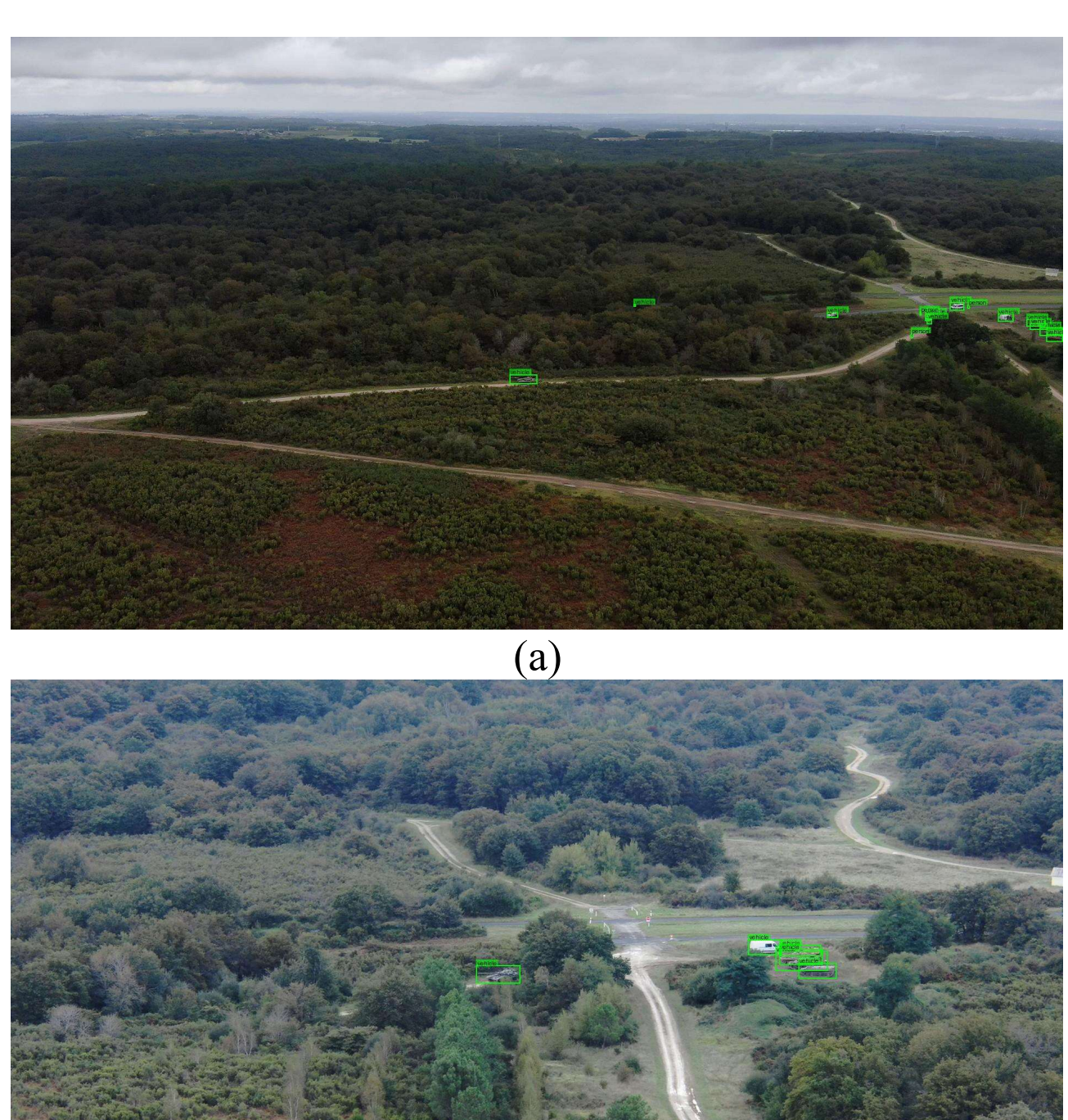

(a)

(b)

Fig. 2. Two examples from the STORE dataset recorded in Fontevraud with an A2G perspective: (a) small objects, (b) medium-sized objects. Images recorded by FlySight S.r.l.

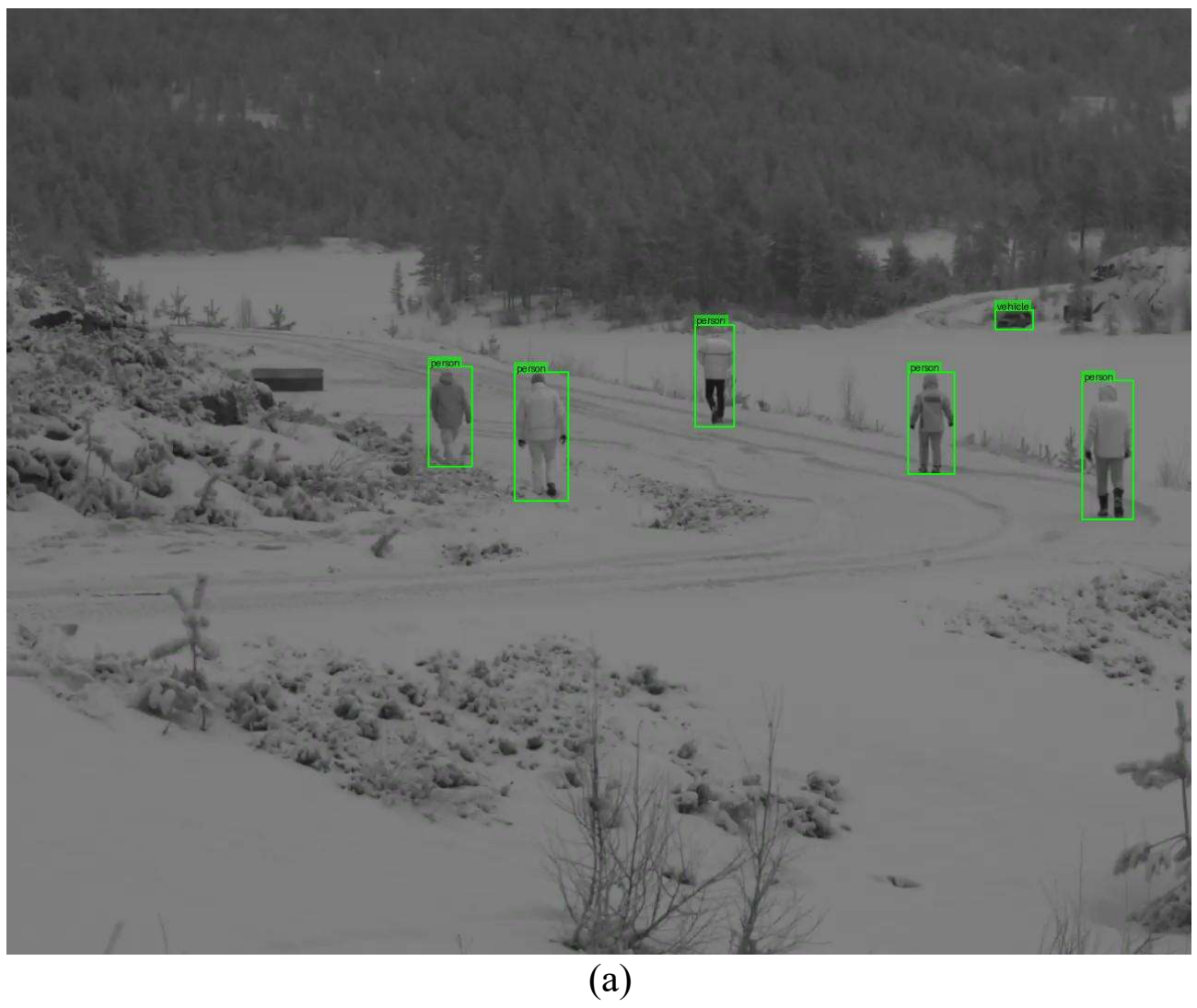

(a)

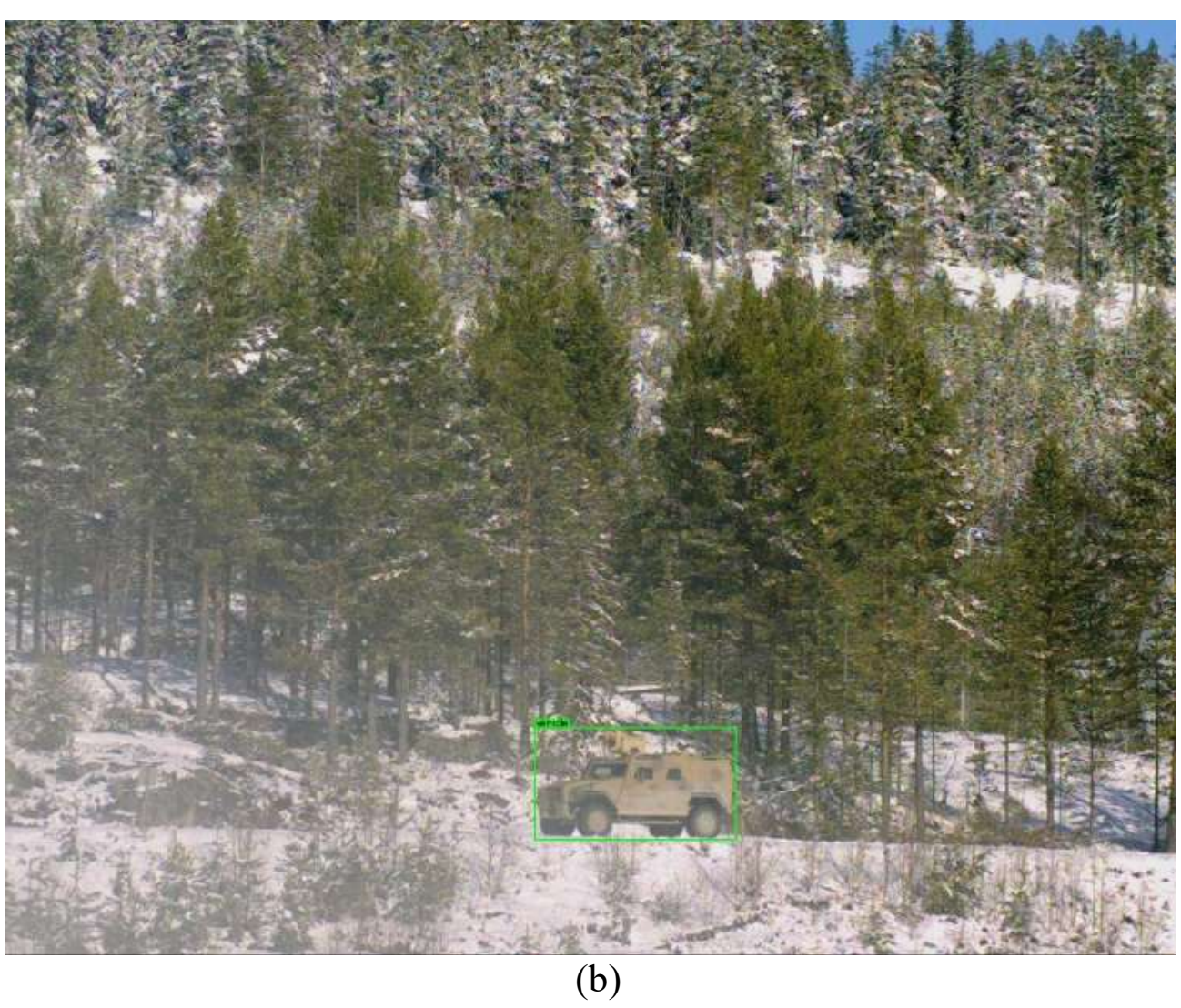

(b)

Fig. 3. Two examples from the STORE dataset recorded near Kongsberg with a G2G perspective: (a) medium-sized objects, (b) large objects. Images recorded by (a) Austrian Institute of Technology GmbH and (b) Safran Electronics and Defense.

The **You Only Look Once (YOLO**) series was originally published in 2016 as a new approach to object detection. Its approach was novel in that it presented a single neural network to predict bounding boxes and class probabilities from full images. This unified architecture made the detection process extremely fast compared to the state-of-the-art at the time, enabling real-time detection [21]. YOLO has since seen several iterations and has become central to the field of object detection [22].

TABLE I. OVERVIEW OF MODELS USED IN THIS STUDY

| Model | Versions | Pretraining data |
|---|---|---|
| YOLOv5 [23, 24] | s, m, l | MS COCO |
| YOLOv8 [25][2] | n, s, m, l, x | MS COCO |
| YOLOv9 [26][3] | t, s, m, c | MS COCO |
| YOLOv10 [27][4] | s, m, l | MS COCO |
| YOLO11 [28][5] | n, s, m, l, x | MS COCO |
| YOLO-World [29, 30][6] | s, m, l | Objects365, GoldG, CC3M-Lite |
| RT-DETR [31, 32][7] | R101, R50, R18 | MS COCO, Object365 |
| D-FINE DETR [33, 34] | n, s, m | MS COCO |

**YOLOv5**, released in 2020, is a one-stage object detection model known for its real-time capabilities, accuracy, and ease of use [24]. Built upon the YOLO framework, it features an improved architecture with CSP (Cross Stage Partial [35])-based backbone for efficient feature extraction, a PANet-inspired neck for multi-scale feature fusion, and a YOLO detection head for precise object localization. YOLOv5 also replaces the SPP (Spatial Pyramid Pooling [36]) block in the backbone with SPPF, an optimized variant that maintains mathematical equivalence while doubling processing speed. Unlike its predecessors, YOLOv5 is implemented in PyTorch, making it highly flexible and easy to fine-tune [23].

**YOLOv8** adopted an anchor-free split head and claimed to improve accuracy and enable more efficient detection compared to anchor-based methods. The model is optimized for balance between accuracy and speed such that it is appropriate for real-time object detection tasks [25]. The model was released as a Python package for ease of use and designed to cater to different settings and tasks including object detection, instance segmentation and keypoint detection [25].

**YOLOv9** addressed the challenges posed by information loss in deep neural networks. The Information Bottleneck Principle and the innovative use of Reversible Functions are central to its design, ensuring YOLOv9 maintains high efficiency and accuracy [26]. The authors proposed the Programmable Gradient Information (PGI) auxiliary supervision framework, which implements two additional branches for training, in addition to the main branch used for inference. These are the Auxiliary Reversible Branch and Multi-level Auxiliary Branch, which were designed for dealing with problems caused by network deepening and error accumulation of deep supervision. Additionally, Generalized Efficient Layer Aggregation Network (GELAN) was introduced to allow the use of any computational blocks, while taking lightweight, inference speed and accuracy into account. PGI and GELAN were combined in YOLOv9 [26].

**YOLOv10** introduced a novel approach called consistent dual assignments, eliminating the need for non-maximum suppression (NMS) during inference [27]. YOLOv10 combined one-to-many and one-to-one matching approaches by introducing an additional one-to-one head, mirroring the original one-to-many branch's structure and optimization objectives. During training, both heads are jointly optimized, leveraging the richer supervision [27].

**YOLO11** introduced several architectural enhancements to improve accuracy and efficiency of real-time object detection [28]. A key innovation was the C3k2 block (Cross Stage Partial with Kernel Size 2), which replaced the previous C2f block by utilizing two smaller convolutions instead of a single large one, leading to faster processing and greater computational efficiency. The SPPF (Spatial Pyramid Pooling - Fast) module was retained to enhance multi-scale feature extraction, while the newly integrated C2PSA (Convolutional Block with Parallel Spatial Attention) improved the model's focus on key image regions, enhancing detection accuracy for small or occluded objects. YOLO11 also incorporated CBS (Convolution-BatchNorm-SiLU) blocks, which refine feature maps to improve stability and overall performance. The detection head leverages multiple C3k2 blocks to process multi-scale features efficiently, optimizing both parameter efficiency and computational speed [28].

**YOLO-World** was created as an adaptation of the YOLO series for open-vocabulary object detection [37, 29]. This was implemented by exploiting pretraining on large-scale datasets, CLIP [38] for extracting text embeddings from captions and labels and YOLOv8 [25] as backbone for extracting image features. A new Re-parameterizable Vision Language Path Aggregation Network (RepVL-PAN) and region-text contrastive loss were introduced to enhance fusion of information between the image and language domain. During inference, a user-defined vocabulary can be encoded and re-parametrized into the RepVL-PAN network, enabling real-time performance. Thus, the parameter counts reported in the results correspond to the models without the text encoder.

The **DEtection TRansformers (DETR)** framework was originally presented as a method that approaches object detection as a direct set prediction problem [39]. Upon release, DETR outperformed existing baselines in object detection. Though its performance on small objects was generally poor, DETR was an improvement in terms of efficiency. Its attention mechanism enabled efficient use of the context of an object for detection and classification. Several variations on the original DETR model have been designed to counter its original shortcomings and improve its efficiency and performance [34, 40].

**Real-Time (RT-)DETR** provided a real-time capable end-to-end Transformer-based detector [31]. End-to-end

[2] Version 8.2.20
[3] Version 8.3.75
[4] Version 8.3.47
[5] Version 8.3.75
[6] Version 8.3.59
[7] Version 1.0

Transformer-based detectors (DETRs) provide a NMS-free alternative to the YOLO series of detectors. However, their high computational cost makes them unable to take advantage of their NMS-free architecture. RT-DETR however, has a reduced computational cost compared to other DETR versions. This was achieved using a 'hybrid encoder to process multi-scale features by decoupling intra-scale interaction and cross-scale fusion' [31]. To increase its accuracy, RT-DETR was uses an uncertainty-minimal query selection scheme, improving the decoder's initial queries [31]. Fifteen versions/sizes of the model are available but since the model training still has relatively high computational costs, only one large version, one medium-sized version, and one small version were evaluated.

**D-FINE DETR** presented a novel approach for generic object detection that re-formulates the anchor-based regression loss for typical object detectors as iteratively refining probability distributions instead [33]. In contrast to RT-DETR and DETR, D-FINE enhances the detection process through its two key components: Fine-grained Distribution Refinement (FDR) and Global Optimal Localization Self-Distillation (GO-LSD). FDR iteratively refines probability distributions to model localization uncertainties more effectively, while GO-LSD facilitates the transfer of localization knowledge from deeper layers to shallower ones, accelerating convergence and improving overall performance [33]. Due to the computational costs of training D-FINE, only the smallest three versions were used.

## III. Experiments and Results

Experiments were conducted by running inference on the out-of-the-box and fine-tuned models for the full STORE dataset. The results were evaluated by assessing COCO %AP at different IoU thresholds, namely 0.5 and 0.5:0.95. Results were split out into performance for A2G and G2G and across object sizes, using the standard COCO object size categories: small, medium and large[8]. The number of parameters is reported to compare models and to give an indication of their real-time capability, relevant to the military operational context.

### *A. Parameters vs. performance*

Fig. 4 shows the parameter count (in millions) compared to performance for all our models and model sizes. When evaluating the results of the out-of-the-box models a trend is visible across models: a higher parameter count also means better performance. This trend persists for the fine-tuned models though the performance of these models is overall slightly worse. This changes when considering only A2G images, which will be discussed in section B.

For out-of-the-box performance, the DETR-based models and YOLO-World stand out compared to the other models from the YOLO series. The difference between the two model families is stronger for the fine-tuned versions, where the performance of the YOLO-based models decreases after fine-tuning while the performance of the DETR-based models stays relatively constant.

### *B. G2G vs. A2G performance*

The fine-tuned models performed worse than the out-of-the-box versions overall. However, when evaluating performance for A2G scenarios, TABLE II. shows that fine-tuning on VisDrone does result in improved performance. The fine-tuned models consistently outperformed the out-of-the-box models for A2G data across model types. On the contrary, fine-tuning harmed performance in the G2G scenario. The out-of-the-box versions perform better in this scenario, causing them to also perform higher overall since G2G is overrepresented in the dataset.

When evaluating the performance across all models, it can be noted that RT-DETR reached the highest performance across all categories, approaching an AP of 30% for both A2G and G2G.

### *C. Performance across object sizes*

The A2G perspective was, overall, more challenging for the models than the G2G perspective. This is, at least partially, explainable by the fact that A2G imagery features smaller objects. TABLE III. shows the performance of the models across object sizes.

The pattern that was visible for A2G vs. G2G is less strong though still noticeable for small vs. medium and large objects: the VisDrone fine-tuned models performed slightly better than the out-of-the-box versions on small objects. However, there are three exceptions to this rule: YOLO11, YOLO-World and D-FINE DETR. Noticeably, these are also some of the largest models included in this overview.

When evaluating the performance across all models, the results are mixed: RT-DETR performs best overall and for small objects, YOLO-World performs best for medium-sized objects and YOLOv5 takes the highest score for large objects. Again, the best performance for small objects is found in a VisDrone fine-tuned model, but the best performance for larger objects is attained with an out-of-the-box model.

TABLE II. %AP@0.5 ACROSS PERSPECTIVES, ONLY THE BEST PERFORMING VERSION(S) FOR EACH MODEL. BEST RESULTS PER MODEL ARE PRINTED IN BOLT, BEST RESULTS PER CATEGORY ARE UNDERLINED.

| Model | Size | Trained on | Both | A2G | G2G |
|---|---|---|---|---|---|
| YOLOv5 | L | - | **23.6** | 15.4 | **22.6** |
| | | VisDrone | 17.9 | **18.5** | 13.1 |
| YOLOv8 | L | - | 22.7 | 13.4 | 22.8 |
| | | VisDrone | 19.0 | **18.4** | 14.7 |
| | X | - | **24.1** | 13.6 | **25.3** |
| | | VisDrone | 18.6 | 16.5 | 14.7 |
| YOLOv9 | C | - | **22.7** | 13.6 | **23.1** |
| | | VisDrone | 18.7 | **19.1** | 14.3 |
| YOLOv10 | X | - | **22.3** | 13.5 | **22.3** |
| | | VisDrone | 19.2 | **18.8** | 12.9 |
| YOLOv11 | X | - | **24.5** | 13.3 | **26.7** |
| | | VisDrone | 16.9 | **18.8** | 10.6 |
| YOLO-World | X | - | **27.1** | 15.8 | **28.3** |
| | | VisDrone | 20.6 | **19.2** | 14.6 |
| RT-DETR | R50 | - | 27.1 | 17.5 | 26.9 |
| | | VisDrone | 26.7 | **27.1** | 18.9 |
| | R101 | - | **29.3** | 18.7 | **29.7** |
| | | VisDrone | 27.3 | 25.8 | 21.9 |
| D-FINE DETR | M | - | **26.2** | 16.9 | **26.2** |
| | | VisDrone | 25.7 | **23.4** | 19.9 |

[8] Small: 0-32*32 pixels, Medium: 32*32-96*96 pixels, Large: 96*96+ pixels

### D. *Worst and best performing tasks*

Now that some trends in model performance have been discussed, we want to highlight the tasks our models performed best and worst at. The only task for which our models attained an AP@0.5:0.95 of over 60% was the detection of large objects from the person class, regardless of perspective (though performance was better for G2G). All out-of-the-box models attained a performance of at least 58% on this task. For detecting both large persons and vehicles, all out-of-the-box models attained an AP of at least 41%.

The models struggled most with small objects in an A2G scenario. Not a single model was able to reach an AP@0.5:0.95 of over 10% in that scenario, though fine-tuned models and versions of RT-DETR performed slightly better than out-of-the-box models in this scenario.

TABLE III. %AP@0.5:0.95 ACROSS OBJECT SIZES, ONLY THE BEST PERFORMING VERSION(S) FOR EACH MODEL. BEST RESULTS PER MODEL ARE PRINTED IN BOLT, BEST RESULTS PER CATEGORY ARE UNDERLINED.

| Model | Size | Trained on | All | S | M | L |
|---|---|---|---|---|---|---|
| YOLOv5 | L | - | **13.1** | 3.2 | **33.8** | **58.9** |
| | | VisDrone | 9.5 | **4.9** | 23.7 | 24.0 |
| YOLOv8 | L | - | 13.4 | 5.7 | 23.7 | 50.4 |
| | | VisDrone | 10.7 | **7.0** | 19.4 | 22.6 |
| | X | - | **14.1** | 6.4 | **23.8** | **52.9** |
| | | VisDrone | 10.8 | 6.5 | 19.6 | 22.3 |
| YOLOv9 | C | - | **13.4** | 6.2 | **22.6** | **50.4** |
| | | VisDrone | 10.4 | **6.7** | 18.8 | 22.6 |
| YOLOv10 | X | - | **13.4** | 5.8 | **23.7** | **50.9** |
| | | VisDrone | 10.6 | **6.7** | 18.8 | 23.0 |
| YOLOv11 | X | - | **14.8** | **7.4** | **24.2** | **53.4** |
| | | VisDrone | 9.1 | 6.1 | 15.9 | 16.1 |
| YOLO-World | X | - | **15.8** | **7.9** | **25.4** | **55.2** |
| | | VisDrone | 11.4 | 6.7 | 19.9 | 33.0 |
| RT-DETR | R101 | - | **15.9** | 7.7 | **25.0** | **54.9** |
| | | VisDrone | 13.5 | **8.1** | 19.8 | 39.7 |
| D-FINE DETR | M | - | **13.8** | **6.6** | **23.2** | **50.9** |
| | | VisDrone | 12.1 | **6.6** | 19.8 | 36.1 |

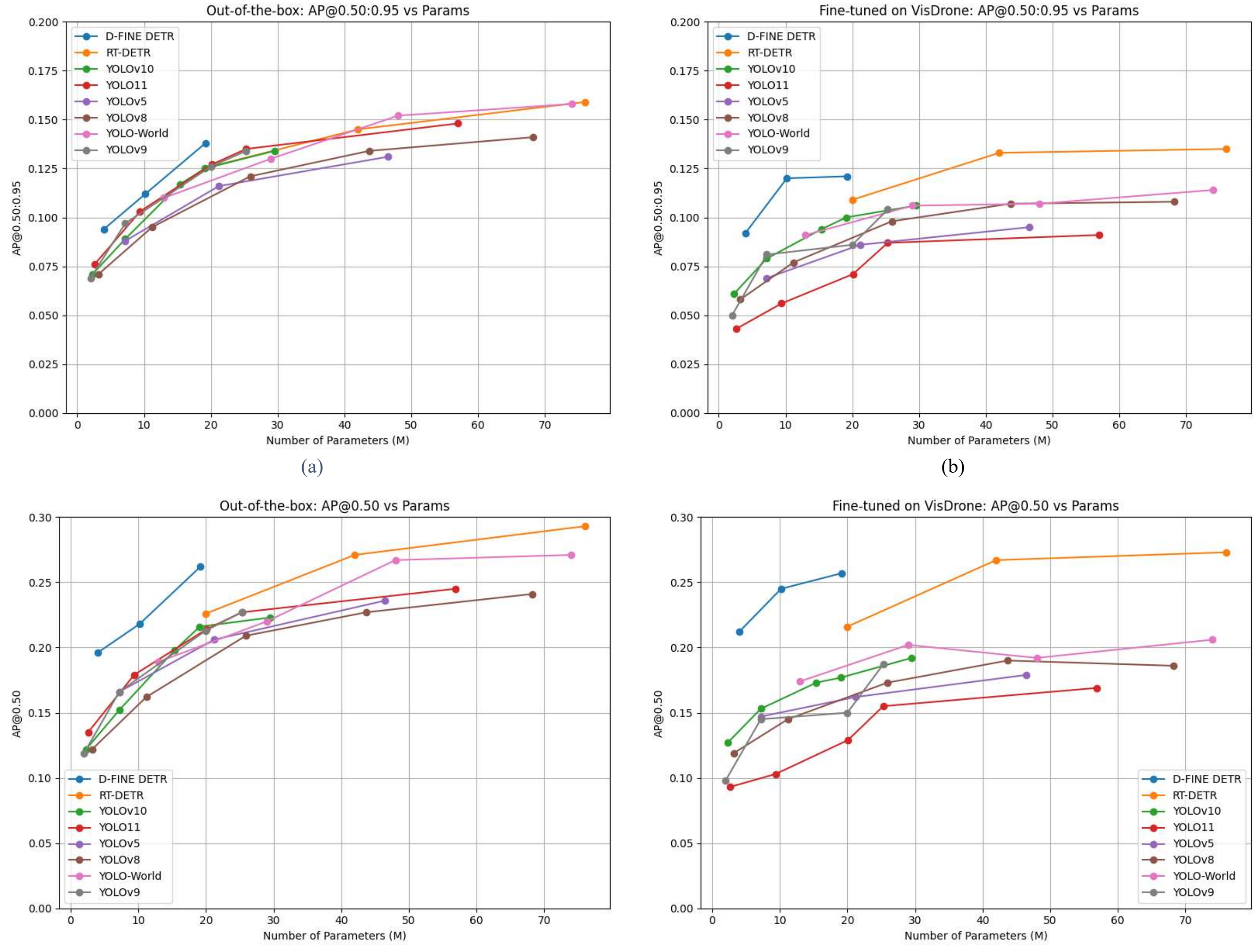

Fig. 4. Several plots showing the number of parameters of the model plotted against performance metrics: (a) AP@0.5:0.95 for the out-of-the-box models, (b) AP@0.5:0.95 for all models fine-tuned on VisDrone (c) AP@0.5 for the out-of-the-box models and (d) AP@0.5 for all models fine-tuned on VisDrone[9]

[9] Due to inconsistent results, YOLOv10 l was excluded from these plots.

## IV. Conclusion and Discussion

In this work, we evaluated the performance of several versions of six iterations of the YOLO series and two variations of the DETR framework. We evaluated the out-of-the-box performance of these models and fine-tuned them on VisDrone, an A2G dataset from the civilian domain. We evaluated their performance on a military, operational dataset to assess if open-source models and datasets could attain reasonable performance in that context.

We found that: (1) bigger models outperform smaller models in this context, (2) DETR-based models show promising results compared to the YOLO series, especially for fine-tuned models, (3) fine-tuning models on an A2G dataset with small objects, even when it is from another domain, improves performance in that setting and, to a lesser extent, for those object sizes, however (4) all models still struggle with detecting small objects in an A2G scenario. Finally, we conclude that, despite recent advances in object detection, in-domain training is still crucial for creating capable ATD/R systems.

### A. Discussion

The two DETR-based models and YOLO-World performed relatively well in our experiments. Especially upon fine-tuning with VisDrone, RT-DETR and D-FINE DETR outperformed the YOLO series. This might be related to the higher generalization ability of the DETR framework [39] as well as the fact that YOLO-World and RT-DETR were trained on more, and different, datasets. However, D-FINE DETR was trained solely on MS COCO, like the YOLO series. This model thus seems to have great potential for the military ATD/R task.

Despite leading to modest improvements in performance in an A2G scenario, fine-tuning on VisDrone harmed performance in the G2G scenario. This could be caused by the out-of-domain nature of VisDrone, being a civilian dataset, as well as the inherent G2G detection capability of many of the pretrained model, since the datasets they are trained on also contain mostly G2G imagery.

Throughout this paper, performance of object detection models in a military context was thoroughly discussed. However, the question of what is 'reasonable performance' in a military, operational context, is in itself an open question to which the answer is highly context-dependent.

Building on this, poor performance shown in this evaluation was occasionally caused by a misalignment of annotations at the barrel of the military vehicle. An example of this is shown in Fig. 5. The ground truth annotation includes the barrel but the model prediction does not. Though this limits the IoU between ground truth and prediction, one can wonder whether this degree of misalignment is truly relevant in an operational context, where the fact that the vehicle was detected at all is most relevant. Additionally, it is not surprising that a model trained on a civilian dataset like VisDrone, which does not include tanks, does not include a barrel into its prediction.

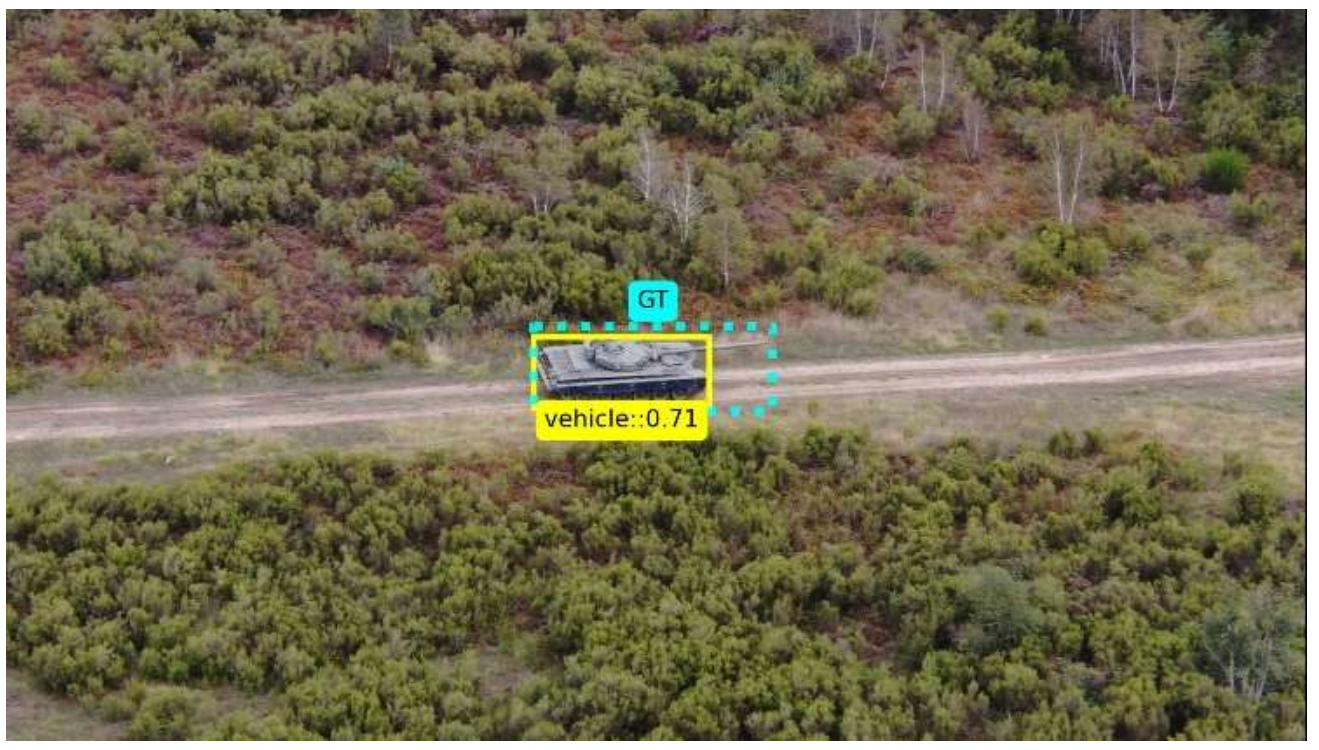

Fig. 5. An example of misalignment between the ground truth annotation (1) and predicted bounding box ('vehicle'). Image recorded by Flysight S.r.l.

Despite these nuances and the promising results for some models and tasks, we conclude that, even when using larger pretraining datasets and superior architectures like DETR, in-domain training still is a crucial part of developing military ATD/R systems.

### B. Limitations

As mentioned previously, not all model versions were included in this evaluation. Particularly, the larger versions of D-FINE DETR were excluded due to computational costs of training. Due to the promising results of this model, this is unfortunate. However, despite this exclusion, we were still able to demonstrate the added value of this model.

Secondly, the used selection of the STORE dataset is not well-balanced. As mentioned earlier, all A2G images included in this study were recorded in Fontevraud, whereas all G2G imagery was recorded in Kongsberg. A mix of both perspectives across locations would have been preferable to enable more robust results and an ablation study of perspective and location.

Thirdly, in combining the VisDrone and STORE datasets, we were forced to aggregate fine-grained (military) vehicle classes into the coarse 'vehicle' class. This limits the applicability of the models presented here, since a human operator would be required to make an assessment of the threat level associated with the vehicle (and/or person) detected by the model. This limitation is accepted since the models were not intended to be used in an operational context but instead serve to show the potential of out-of-the-box state-of-the-art methods.

Lastly, the inclusion of a wider set of metrics, including those using a lower IoU threshold, as discussed in section A, might have shown higher and more operationally relevant performance. For the sake of simplicity and to give an indication of the feasibility of using open-source models and datasets in a military use case, the current metrics still provide a comprehensive overview of the potential of the studied models. Another set of metrics that could have been included is those representing real-time capability and computational performance more accurately, such as FLOPS and latency. Due to the collaborative nature of this evaluation effort, this was not feasible and parameter size was presented instead to give an indication of the computational complexity of each model.

### C. Future work

As shown in this evaluation, there is still value in fine-tuning models on in-domain data to ensure reasonable performance in a military operational context, particularly in A2G scenarios and when facing small objects. We aim to incorporate in-domain training in future studies, specifically within EDF STORE. Within that context, it will be particularly worthwhile to fine-tune DETR-based models and YOLO-World, but most prominently D-FINE DETR and its larger versions. An underexplored direction that could be considered in future work is the use of out-of-the-box Vision-Language Models, as has already been done for fine-grained classification [41, 4].

When conducting such in-domain training, it will be valuable to perform similar evaluations as the one presented here, to highlight the performance gains enabled by specific training strategies. Inclusion of a wider and more representative set of metrics that align with the goals of military ATD/R, and particularly small object detection, will be worthwhile in such an evaluation. Van Leeuwen et al. [42] make some initial suggestions for such metrics.

In terms of training on in-domain data, a final suggestion for future work is to generate and train on synthetic data, as has also been previously successively attempted, especially in combination with a Transformer backbone [5, 6]. Though the suggestion of in-domain training with real data seems obvious, acting on it is less than straightforward. Datasets, such as the STORE dataset, are still limited in availability. We highly recommend investing in the creation of such datasets with diverse contexts, perspectives, occlusion rates and objects, to develop capable ATD/R systems.

## Acknowledgment

This work received funding from the European Defence Fund through the project STORE (Shared daTabase for Optronics image Recognition and Evaluation), grant agreement №101121405. We would like to particularly express our gratitude to all consortium partners involved in the data acquisition and processing.

## References

[1] Z.-Q. Zhao, P. Zheng, S.-t. Xu and X. Wu, "Object detection with deep learning: A review," *IEEE transactions on neural networks and learning systems,* vol. 30, no. 11, pp. 3212--3232, 2019.

[2] A. R. Pathak, M. Pandey and S. Rautaray, "Application of deep learning for object detection," *Procedia computer science,* vol. 132, pp. 1706-1717, 2018.

[3] M. A. Munir, M. H. Khan, M. Sarfraz and M. Ali, "Towards improving calibration in object detection under domain shift," *Advances in Neural Information Processing Systems,* vol. 35, pp. 38706-38718, 2022.

[4] F. G. Heslinga, T. A. Eker, E. P. Fokkinga, J. E. van Woerden, F. A. Ruis, R. J. den Hollander and K. Schutte, "Combining simulated data, foundation models, and few real samples for training object detectors," in *Synthetic Data for Artificial Intelligence and Machine Learning: Tools, Techniques, and Applications II*, SPIE, 2024, pp. 70-81.

[5] F. A. Ruis, A. M. Liezenga, F. G. Heslinga, L. Ballan, T. A. Eker, R. J. den Hollander, M. C. van Leeuwen and J. a. H. W. Dijk, "Improving object detector training on synthetic data by starting with a strong baseline methodology," in *Synthetic Data for Artificial Intelligence and Machine Learning: Tools, Techniques, and Applications II*, 2024, pp. 358-370}.

[6] F. G. Heslinga, E. P. Fokkinga, T. H. Eker, A. M. Liezenga, R. J. den Hollander, V. O. Oppeneer, A. van Heteren, R. van Vossen, H. J. Kuijf, J. J. van de Sande and others, "On the use of simulated data for target recognition and mission planning," in *Artificial Intelligence for Security and Defence Applications II*, SPIE, 2024, pp. 209-231.

[7] A. Bulat, R. Guerrero, B. Martinez and G. Tzimiropoulos, "FS-DETR: Few-Shot DEtection TRansformer with Prompting and without Re-Training," in *IEEE/CVF international conference on computer vision*, 2023.

[8] Y. Zhao, W. Lv, S. Xu, J. Wei, G. Wang, Q. Dang, Y. Liu and J. Chen, "DETRs Beat YOLOs on Real-time Object Detection," in *IEEE/CVF conference on computer vision and pattern recognition*, 2024.

[9] M. L. Ali and Z. Zhang, "The YOLO framework: A comprehensive review of evolution, applications, and benchmarks in object detection," *Computers,* vol. 13, no. 12, p. 336, 2024.

[10] T. Ren, S. Liu, F. Li, H. Zhang, A. Zeng, J. Yang, X. Liao, D. Jia, H. Li, H. Cao and others, "detrex: Benchmarking detection transformers," *arXiv preprint arXiv:2306.07265,* 2023.

[11] P. Zhu, L. Wen, X. Bian, H. Ling and Q. Hu, "Vision meets drones: A challenge," *arXiv preprint arXiv:1804.07437,* 2018.

[12] T.-Y. Lin, M. Maire, S. Belongie, J. Hays, P. Perona, D. Ramanan, P. Dollár and C. L. Zitnick, "Microsoft coco: Common objects in context," in *Computer vision--ECCV 2014: 13th European conference, zurich, Switzerland, September 6-12, 2014, proceedings, part v 13*, 2014.

[13] S. Shao, Z. Li, T. Zhang, C. Peng, G. Yu, X. Zhang, J. Li and J. Sun, "Objects365: A large-scale, high-quality dataset for object detection," in *Proceedings of the IEEE/CVF international conference on computer vision*, 2019.

[14] A. Kamath, M. Singh, Y. LeCun, G. Synnaeve, I. Misra and N. Carion, "Mdetr-modulated detection for end-to-end multi-modal understanding," in *IEEE/CVF international conference on computer vision*, 2021.

[15] P. Sharma, N. Ding, S. Goodman and R. Soricut, "Conceptual captions: A cleaned, hypernymed, image alt-text dataset for automatic image captioning," in *Proceedings of the 56th Annual Meeting of the Association for Computational Linguistics (Volume 1: Long Papers)*, 2018, pp. 2556-2565.

[16] B. A. Plummer, L. Wang, C. M. Cervantes, J. C. Caicedo, J. Hockenmaier and S. Lazebnik, "Flickr30k entities: Collecting region-to-phrase correspondences for richer image-to-sentence models," in *Proceedings of the IEEE international conference on computer vision*, 2015.

[17] D. A. Hudson and C. D. Manning, "Gqa: A new dataset for real-world visual reasoning and compositional question answering," in *Proceedings of the IEEE/CVF conference on computer vision and pattern recognition*, 2019.

[18] Y. Cao, Z. He, L. Wang, W. Wang, Y. Yuan, D. Zhang, J. Zhang, P. Zhu, L. van Gool, J. Han, S. Hoi, Q. Hu and M. Liu, "VisDrone-DET2021: The Vision Meets Drone Object Detection Challenge Results," in *Proceedings of the IEEE/CVF International Conference on Computer Vision (ICCV) Workshops*, 2021.

[19] D. Du, P. Zhu, L. Wen, X. Bian, H. Lin, Q. Hu, T. Peng, J. Zheng, X. Wang, Y. Zhang and others, "VisDrone-DET2019: The vision meets drone object detection in image challenge results," in *Proceedings of the IEEE/CVF international conference on computer vision workshops*, 2019.

[20] S. Langlois, W. Achour, S. Bensberg, N. Camarlinghi, M. Ehrhart, N. Klier, A. Lemeur, S. Legoupil, A. Liezenga, R. Pflugfelder, T. Pruuden and K.-D. Saasen, "Challenges in creating an experimental database for artificial intelligence: the EDF STORE project," in *12th international symposium on optronics in defence & security*, Marseille, 2026.

[21] J. Redmon, S. Divvala, R. Girshick and A. Farhadi, "You only look once: Unified, real-time object detection," in *Proceedings of the IEEE conference on computer vision and pattern recognition*, 2016.

[22] J. Terven, D.-M. Córdova-Esparza and J.-A. Romero-González, "A comprehensive review of yolo architectures in computer vision: From

yolov1 to yolov8 and yolo-nas," *Machine learning and knowledge extraction,* vol. 5, no. 4, pp. 1680--1716, 2023.

[23] G. Jocher and others, "YOLOv5," Ultralytics, 25 June 2020. [Online]. Available: https://github.com/ultralytics/yolov5. [Accessed 15 August 2025].

[24] G. Jocher, "Comprehensive Guide to Ultralytics YOLOv5," Ultralytics, 12 November 2023. [Online]. Available: https://docs.ultralytics.com/yolov5/. [Accessed 21 March 2025].

[25] G. Jocher and e. al., "Ultralytics Yolov8 documentation," Ultralytics, 12 November 2023. [Online]. Available: https://docs.ultralytics.com/models/yolov8. [Accessed 28 February 2025].

[26] C.-Y. Wang, I.-H. Yeh and H.-Y. Mark Liao, "Yolov9: Learning what you want to learn using programmable gradient information," in *European conference on computer vision*, Springer, 2024, pp. 1-21.

[27] A. Wang, H. Chen, L. Liu, K. CHEN, Z. Lin, J. Han and G. Ding, "YOLOv10: Real-Time End-to-End Object Detection," in *The Thirty-eighth Annual Conference on Neural Information Processing Systems*, 2024.

[28] R. Khanam and M. Hussain, "YOLOv11: An overview of the key architectural enhancements," *arXiv preprint arXiv:2410.17725,* 2024.

[29] T. Cheng, L. Song, Y. Ge, W. Liu, X. Wang and Y. Shan, "YOLO-World: Real-Time Open-Vocabulary Object Detection," in *IEEE/CVF conference on computer vision and pattern recognition*, 2024.

[30] G. Jocher and others, "YOLO-World Model," Ultralytics, 14 February 2024. [Online]. Available: https://docs.ultralytics.com/models/yolo-world/.

[31] Y. Zhao, W. Lv, S. Xu, J. Wei, G. Wang, Q. Dang, Y. Liu and J. Chen, "DETRs Beat YOLOs on Real-time Object Detection," in *The IEEE/CVF Conference on Computer Vision and Pattern Recognition 2024*, 2024.

[32] Y. Zhao, W. Lv, S. Xu, J. Wei, G. Wang, Q. Dang, Y. Liu and J. Chen, "RT-DETR: DETRs Beat YOLOs on Real-time Object Detection," 17 April 2023. [Online]. Available: https://github.com/lyuwenyu/RT-DETR.

[33] Y. Peng, H. Li, P. Wu, Y. Zhang, X. Sun and F. Wu, "D-FINE: Redefine Regression Task in DETRs as Fine-grained Distribution Refinement," *arXiv preprint arXiv:2410.13842,* 2024.

[34] Y. Peng, H. Li, P. Wu, Y. Zhang, X. Sun and F. Wu, "D-FINE: Redefine Regression Task of DETRs as Fine-grained Distribution Refinement," 18 March 2025. [Online]. Available: https://github.com/Peterande/D-FINE/. [Accessed 11 November 2024].

[35] C.-Y. Wang, H.-Y. M. Liao, Y.-H. Wu, P.-Y. Chen, J.-W. Hsieh and I.-H. Yeh, "CSPNet: A new backbone that can enhance learning capability of CNN," in *Proceedings of the IEEE/CVF conference on computer vision and pattern recognition workshops*, 2020.

[36] K. He, X. Zhang, S. Ren and J. Sun, "Spatial pyramid pooling in deep convolutional networks for visual recognition," *IEEE transactions on pattern analysis and machine intelligence,* vol. 37, no. 9, pp. 1904-1916, 2015.

[37] A. Zareian, K. D. Rosa, D. H. Hu and S.-F. Chang, "Open-vocabulary object detection using captions," in *IEEE/CVF conference on computer vision and pattern recognition*, 2021.

[38] A. Radford, J. W. Kim, C. Hallacy, A. Ramesh, G. Goh, S. Agarwal, G. Sastry, A. Askell, P. Mishkin, J. Clark and others, "Learning transferable visual models from natural language supervision," in *International conference on machine learning*, PmLR, 2021, pp. 8748-8763.

[39] N. Carion, F. Massa, G. Synnaeve, N. Usunier, A. Kirillov and S. Zagoruyko, "End-to-end object detection with transformers," *European conference on computer vision,* pp. 213-229, 2020.

[40] W. Chen, J. Luo, F. Zhang and Z. Tian, "A review of object detection: Datasets, performance evaluation, architecture, applications and current trends," *Multimedia Tools and Applications,* vol. 83, no. 24, pp. 65603-65661, 2024.

[41] J. E. van Woerden, G. Burghouts, L. Nijskens, A. M. Liezenga, S. van Rooij, F. Ruis and H. J. Kuijf, "Occlusion robustness of CLIP for military vehicle classification," in *Artificial Intelligence for Security and Defence Applications III*, Madrid, SPIE, 2025, pp. 412-422.

[42] M. C. van Leeuwen, E. P. Fokkinga, W. Huizinga, J. Baan and F. G. Heslinga, "Toward versatile small object detection with Temporal-YOLOv8," *Sensors,* vol. 24, no. 22, p. 7387, 2024.